\documentclass{article}
\usepackage{spconf,amsmath,graphicx}
\usepackage{bm}
\usepackage{amssymb}
\usepackage{booktabs}
\DeclareMathOperator*{\argmax}{argmax}
\newcommand{\tabincell}[2]{\begin{tabular}{@{}#1@{}}#2\end{tabular}}

\title{Neural CRF Transducers for Sequence Labeling}
\name{Kai Hu${}^{\dagger}$, Zhijian Ou${}^{\dagger}$, Min Hu${}^{\ddagger}$, Junlan Feng${}^{\ddagger}$ \thanks{This work is supported by NSFC 61473168, Ministry of Education and China Mobile joint funding MCM20170301. Correspondence to: Z. Ou.}}
\address{${}^{\dagger}$Speech Processing and Machine Intelligence (SPMI) Lab, Tsinghua University, China\\
	${}^{\ddagger}$ China Mobile Research Institute\\
	{huk17@mails.tsinghua.edu.cn, ozj@tsinghua.edu.cn, \{humin,fengjunlan\}@chinamobile.com}
}

\begin{document}
\ninept

\maketitle
\begin{abstract}
Conditional random fields (CRFs) have been shown to be one of the most successful approaches to sequence labeling.
Various linear-chain neural CRFs (NCRFs) are developed to implement the non-linear node potentials in CRFs, but still keeping the linear-chain hidden structure.
In this paper, we propose NCRF transducers, which consists of two RNNs, one extracting features from observations and the other capturing (theoretically infinite) long-range dependencies between labels.
Different sequence labeling methods are evaluated over POS tagging, chunking and NER (English, Dutch).
Experiment results show that NCRF transducers achieve consistent improvements over linear-chain NCRFs and RNN transducers across all the four tasks, and can improve state-of-the-art results.

\end{abstract}
\section{Introduction}
\label{sec:intro}

Given a sequence of observations $x \triangleq x_1,\cdots\,x_n$, the task of sequence labeling is to predict a sequence of labels $y \triangleq y_1,\cdots\,y_n$, with one label for one observation in each position.
Sequence labeling is of broad interest and has been popularly applied in the area of natural language processing (NLP), like in part-of-speech (POS) tagging \cite{collobert2011natural,ling2015finding}, named entity recognition (NER) \cite{huang2015bidirectional,lample2016neural,ma2016end}, chunking \cite{huang2015bidirectional,sogaard2016deep}, syntactic parsing \cite{durrett2015neural} and semantic slot filling \cite{zhai2017neural}, and also in other areas such as bioinformatics \cite{Sato2005RNASS,peng2009conditional}.

Conditional random fields (CRFs) \cite{lafferty2001conditional} have been shown to be one of the most successful approaches to sequence labeling.
A recent progress is to develop Neural CRF (NCRF) models, which combines the sequence-level discriminative ability of CRFs and the representation ability of neural networks (NNs), particularly the recurrent NNs (RNNs). 
These models have achieved state-of-the-art results on a variety of sequence labeling tasks, and in different studies, are called conditional neural field \cite{peng2009conditional}, neural CRF \cite{artieres2010neural}, recurrent CRF \cite{Mesnil2015UsingRN}, and LSTM-CRF \cite{lample2016neural,ma2016end}.
Though there are detailed differences between these existing models, generally they are all defined by using NNs (of different network architectures) to implement the non-linear node potentials in CRFs, while still keeping the linear-chain hidden structure (i.e. using a bigram table as the edge potential).
For convenience, we refer to these existing combinations of CRFs and NNs as \emph{linear-chain NCRFs} in general.
This represents an extension from conventional CRFs, where both node potentials and edge potentials are implemented as linear functions using discrete indicator features.

In this paper, we present a further extension and propose \emph{neural CRF transducers}, which introduce a LSTM-RNN to implement a new edge potential so that long-range dependencies in the label sequence are captured and modeled.
In contrast, linear-chain NCRFs capture only first-order interactions and neglect higher-order dependencies between labels, which can be potentially useful in real-world sequence labeling applications, e.g. as shown in \cite{zhang2018does} for chunking and NER.

There are two LSTM-RNNs in a NCRF transducer, one extracting features from observations and the other capturing (theoretically infinite) long-range dependencies between labels.
In this view, a NCRF transducer is similar to a RNN transducer \cite{graves2012sequence}, which also uses two LSTM-RNNs.
Additionally, the recent attention-based seq2seq models \cite{bahdanau2014neural} also use an LSTM-based decoder to exploit long-range dependences between labels. 
However, both RNN transducers and seq2seq models, as locally normalized models, produce position-by-position conditional distributions
over output labels, and thus suffer from the label
bias and exposure bias problems \cite{lafferty2001conditional,andor2016globally,wiseman2016sequence}.
In contrast, NCRF transducers are globally normalized, which overcome these two problems.
We leave more discussions about existing related studies to section \ref{sec:related work}.

Different sequence labeling methods are evaluated over POS tagging, chunking and NER (English, Dutch).
Experiment results show that NCRF transducers achieve consistent improvements over linear-chain NCRFs and RNN transducers across all the four tasks.
Notably, in the CoNLL-2003 English NER task, the NCRF transducer achieves state-of-the-art $F_1$ (92.36), better than 92.22  \cite{peters2018deep}, using the same deep contextualized word representations.

\section{Background}
\label{sec:background}

\textbf{Linear-chain NCRFs.}	A linear-chain CRF defines 
a conditional distribution for label sequence $y$ given observation sequence $x$ :
\begin{displaymath}
p(y|x) \propto \exp \left\lbrace   \sum_{i=1}^n\phi_i(y_i,x)+
\sum_{i=1}^n \psi_i(y_{i-1},y_i,x) \right\rbrace.
\end{displaymath}
Here the labels $y_i$'s are structured to form a chain, giving the term linear-chain. 
$\phi_i(y_i,x)$ is the node potential defined at position $i$, which, in recently developed NCRFs \cite{collobert2011natural,huang2015bidirectional,lample2016neural,ma2016end,peng2009conditional} is implemented by using features generated from a NN of different network architectures.
$\psi_i(y_{i-1},y_i,x)$ is the edge potential defined on the edge connecting $y_{i-1}$ and $y_i$, which, in these existing NCRFs, is mostly implemented as a transition matrix $A$:
\begin{displaymath}
\psi_i(y_{i-1}=j,y_i=k,x) = A_{j,k}
\end{displaymath}
This edge potentials admit efficient algorithms for training and decoding, but only capture first-order dependencies between labels.

\textbf{RNN Transducers} are originally developed for general sequence-to-sequence learning \cite{graves2012sequence}, which do not assume that the input and output sequences are of equal lengths and aligned, e.g. in speech recognition. However, it can be easily seen that RNN transducers can be applied to sequence labeling as well, by defining  
$ p(y|x)=\prod_{i=1}^n p(y_i|y_{0:i-1},x)$ and implementing $p(y_i|y_{0:i-1},x)$ through two networks - transcription network $F$ and prediction network $G$ as follows:
\begin{equation} \label{eq:RNNT-conditional}
p(y_i=k|y_{0:i-1},x) = \frac{\exp({f_i^k + g_i^k})}{\sum_{k'=1}^{K}\exp({f_i^{k'} + g_i^{k'}})}
\end{equation}
Here $F$ scans the observation sequence $x$ and outputs the transcription vector sequence $f \triangleq f_1,\cdots\,f_n$. 
$G$ scans the label sequence $y_{0:n-1}$ and outputs the prediction vector sequence $g \triangleq g_1,\cdots\,g_n$. 
$y_0$ denotes the beginning symbol ($<bos>$) of the label sequence. 
For a sequence labeling task with $K$ possible labels, $f_i$ and $g_i$ are $K$ dimensional vectors.
Superscript $k$ is used to denote the $k^{th}$ element of the vectors.
Remarkably, the prediction network $G$ can be viewed as a label language model, capable of modeling long-range dependencies in $y$, which is exactly the motivation to introducing $G$ in RNN transducers.

To ease comparison, we will also refer to the network below the CRF layer in linear-chain NCRFs as a transcription network, since we also implement $\phi_i(y_i=k,x;\theta)$ as $f_i^k$ in our experiments.

\section{NCRF Transducers}
\label{sec:CRFT}

\begin{table}[t]
	\begin{footnotesize}
		\begin{center}
			\caption{\small{Model comparison and connection.}}
			\label{table:model-table}
			\begin{tabular}{lcc}
			\toprule
		\multicolumn{1}{l}{\textbf{Model}}  &\multicolumn{1}{c}{\tabincell{c}{\textbf{Globally}  \\ \textbf{normalized}}} &\multicolumn{1}{c}{\tabincell{c}{\textbf{Long-range dependencies} \\ \textbf{between labels}} } \\
\midrule
			Linear-chain NCRF  &$\surd$ &$\times$ \\
			RNN Transducer  &$\times$ &$\surd$ \\
			NCRF Transducer &$\surd$ &$\surd$ \\
\bottomrule
			\end{tabular}
		\end{center}	
	\end{footnotesize}
\vspace{-0.5cm}
\end{table}

In the following, we develop NCRF transducers, which combine the advantages of linear-chain NCRFs (globally normalized, using LSTM-RNNs to implenent node potentials) and of RNN transducers (capable of capturing long-range dependencies in labels), and meanwhile overcome their drawbacks, as illustrated in Table \ref{table:model-table}.

\subsection{Model definition}
\label{ssec:Model description}

A NCRF transducer defines a globally normalized, conditional distribution $p(y|x;\theta)$ as follows:
\begin{displaymath}
\ p(y|x;\theta) = \frac{\exp \left\lbrace  u(y,x;\theta)\right\rbrace }{Z(x;\theta)}.
\end{displaymath}
where $Z(x;\theta) = \sum_{y'\in\mathcal{D}_n}\exp \left\lbrace u(y',x;\theta) \right\rbrace $ is the global normalizing term and $\mathcal{D}_n$ is the set of allowed label sequences of length $n$. The total potential $u(y,x;\theta)$ is decomposed as follows:
\begin{displaymath}
u(y,x;\theta) = \sum_{i=1}^{n} \left\lbrace \phi_i(y_i,x;\theta) + \psi_i(y_{0:i-1},y_i;\theta) \right\rbrace.
\end{displaymath}
where $\phi_i(y_i,x;\theta)$ is the node potential at position $i$, $\psi_i(y_{0:i-1},y_i;\theta)$ is the clique potential involving labels from the beginning up to position $i$. Thus the underlying undirected graph for the label sequence $y$ is fully-connected, which potentially can capture long-range dependencies from the beginning up to each current position.

\begin{figure}[htb]
	\centering
	\centerline{\includegraphics[width=8.5cm]{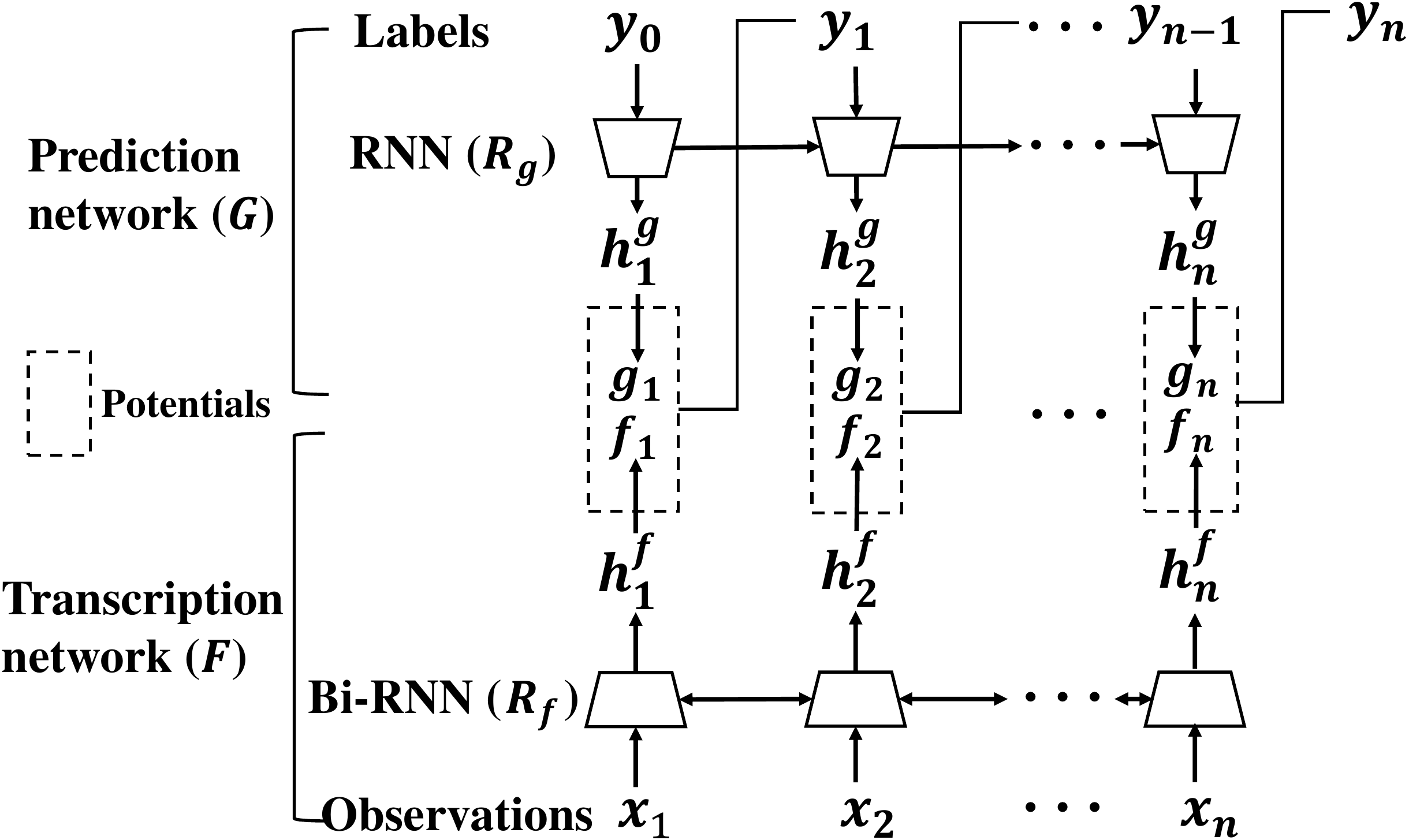}}
	\caption{The architecture of a NCRF transducer.}
	\label{fig:crft}
\end{figure}

\subsection{Neural network architectures}
\label{ssec:Neural network architecture}

Like in RNN transducers, we introduce two networks in NCRF transducers, as shown in Fig. \ref{fig:crft}. The transcription network $F$ implements the node potential $\phi_i(y_i,x;\theta)$, which represents the score for $y_i$ based on observations $x$. In our experiments on NLP sequence labeling, each word $x_i$ is represented by a concatenation of a pretrained word embedding vector and another embedding vector obtained from a character-level CNN.
The transcription network $F$ is a bidirectional
RNN ($R_f$) that scans the sequence of the concatenated vectors for words to generate hidden vectors $h_i^f=[\overrightarrow{h_i^f};\overleftarrow{h_i^f}]$, which are then fed to a linear layer with output size of $K$ to generate $f_i \in \mathbb{R}^K$.

The prediction network $G$ implements the clique potential $\psi_i(y_{0:i-1},y_i;\theta)$, which represents the score for $y_i$ by taking account of dependencies between $y_i$ and previous labels $y_{0:i-1}$.
In our experiments, each label $y_i$ is represented by a label embedding vector, initialized randomly. $G$ is a unidirectional RNN ($R_g$) that accepts the label sequence $y$ and generates hidden vectors $h_i^g=\overrightarrow{h_i^g}$, which are then fed to a linear layer with output size of $K$ to generate $g_i \in \mathbb{R}^K$.

It can be seen from above that a NCRF transducer is similar to a RNN transducer. The difference is that a RNN transducer is local normalized through softmax calculations as shown in Eq. (\ref{eq:RNNT-conditional}), while a NCRF transducer is globally normalized, locally producing (un-normalized) potential scores.

\subsection{Potential design}
\label{ssec:potentials design}

Based on $f_i$ and $g_i$, there are two possible designs to implement the potentials $\phi_i$ and $\psi_i$, which are chosen empirically in our experiments. The first design is:
\begin{equation} \label{eq:design-a}
\begin{aligned}
\phi_i(y_i=k,x;\theta)= f_i^k \\
\psi_i(y_{0:i-1},y_i=k;\theta)= g_i^k
\end{aligned}
\end{equation}
The second design is:
\begin{equation} \label{eq:design-b}
\begin{aligned}
\phi_i(y_i=k,x;\theta) = log \frac{\exp({f_i^k})}{\sum_{k'=1}^{K}\exp(f_i^{k'})} \\
\psi_i(y_{0:i-1},y_i=k;\theta)= log \frac{\exp({g_i^k})}{\sum_{k'=1}^{K}\exp(g_i^{k'})}
\end{aligned}
\end{equation}

\subsection{Decoding and training}
\label{ssec:training}

NCRF transducers break the first-order Markov assumption in the label sequence as in linear-chain NCRFs and thus do not admit dynamic programming for decoding. Instead, we use beam search to approximately find the most probable label sequence:
\begin{displaymath}
\hat{y}=\argmax_{y'\in\mathcal{D}_n} p(y'|x;\theta) = \argmax_{y'\in\mathcal{D}_n} u(y',x;\theta).
\end{displaymath}

Training data consists of inputs $x$ paired with oracle
label sequences $y^*$. We use stochastic gradient
descent (SGD) on the negative log-likelihood of the
training data:
\begin{displaymath}
L(y^*;\theta) = -u(y^*,x;\theta) + \log Z(x;\theta).
\end{displaymath}
It is easy to calculate the gradient of the first term. However, the gradient of the log normalizing term involves model expectation:
\begin{displaymath}
\nabla_\theta \log Z(x;\theta) = E_{p(y'|x;\theta)} \left[ \nabla_\theta u(y',x;\theta) \right] 
\end{displaymath}
The calculations of the normalizing term and the model expectation can be exactly performed for linear-chain NCRFs (via the forward and backward algorithm), but are intractable for NCRF transducers.
It is empirically found in our experiments that the method of beam search with early updates \cite{collins2004incremental} marginally outperforms Monte Carlo based methods for training NCRF transducers.

The basic idea is that we run beam search and approximate the normalizing term by summing over the paths in the beam.
Early updates refer to that as the training sequence is being decoded, we keep track of the location of the oracle path in the beam; If the oracle path falls out of the beam at step $j$, a stochastic gradient step is taken on the following objective:
\begin{displaymath}
L(y^*_{1:j};\theta) = -u(y^*_{1:j};\theta) + log \sum_{y'\in\mathcal{B}_j}\exp \left\lbrace u(y'_{1:j};\theta) \right\rbrace
\end{displaymath}
where $u(y_{1:j};\theta) = \sum_{i=1}^{j} \left\lbrace \phi_i(y_i,x;\theta) + \psi_i(y_{0:i-1},y_i;\theta) \right\rbrace $ denotes the partial potential (with abuse of the notation of $u$).
The set $\mathcal{B}_j$ contains all paths in the beam
at step $j$, together with the oracle path prefix $y^*_{1:j}$.

\section{Experimental setup}
\label{sec:experiments setup}

Different sequence labeling methods are evaluated over four tasks - POS tagging, chunking and NER (English, Dutch).
We replace all digits with zeros and rare words (frequency less than 1) by $<$UNK$>$, as a common pre-processing step for all methods.

\textbf{Datasets.}
The following benchmark datasets are used - PTB POS tagging, CoNLL-2000 chunking, CoNLL-2003 English NER and CoNLL-2002 Dutch NER. For the task of POS tagging, we follow the previous work \cite{ma2016end} to split the dataset and report accuracy. For the NER tasks, we follow the previous work \cite{ma2016end} to use the BIOES tagging scheme and report micro-average $F_1$. For the chunking task, we follow the previous work \cite{peters2017semi}; 1000 sentences are randomly sampled from the training set to be the development set. BIOES tagging scheme and $F_1$ are used.

\textbf{Model configuration.}
For word embeddings, 100-dim Glove embeddings \cite{pennington2014glove} are used for the tasks of POS tagging and English NER. 64-dim skip-n-gram embeddings \cite{lample2016neural} are used for the Dutch NER task. 50-dim Senna embeddings \cite{collobert2011natural} are used for the chunking task. 
For character embeddings, we use 30-dim embeddings (randomly initialized) and a CNN consisting of 30 filters with 3-character width. For label embeddings, 10-dim embeddings are used (randomly initialized).

For the transcription network $F$, a bidirectional LSTM, consisting of one layer of 200 hidden units, is used in the POS tagging and NER tasks; the bidirectional LSTM used in the chunking task consists of two layers with 200 hidden units each layer. 
For the prediction network $G$, an unidirectional LSTM, consisting of one layer of 50 hidden units, is used in all tasks.

For optimizers, SGD with momentum 0.9 is used in the tasks of POS tagging and English NER, the initial learning rate is 0.01 with a 0.05 decay rate per epoch as in the previous work \cite{ma2016end}. 
In the tasks of chunking and Dutch NER, Adam is used, the initial learning rate being fixed at 1e-3 without decay.
Beam width 128 is used for training and 512 is used for decoding.
The mini-batch size is set to be 16. 
Following the previous work \cite{ma2016end}, we add 0.5 dropout.

In our experiments, NCRF transducers are initialized with the weights from pre-trained RNN transducers, which is found to yield better and faster learning.
When finetuning the pre-trained model, we set the initial learning rate to be 5e-3 and adopt SGD with momentum 0.9. In each task, we tune hyperparameters on the development set and use early stopping.

\begin{table}[t]
	\begin{footnotesize}
		\begin{center}
			\caption{\small{POS tagging results over PTB dataset.}}
			\label{table:pos}
			\begin{tabular}{lc}
				\toprule
				\multicolumn{1}{l}{\bf Model}  &\multicolumn{1}{c}{\bf Accuracy} \\
				\midrule
				Collobert et al \cite{collobert2011natural}         &97.29 \\
				Manning \cite{manning2011part}             &97.28 \\
				Santos $\&$ Zadrozny \cite{santos2014learning}     &97.32 \\
				Sun \cite{sun2014structure}            &97.36 \\
				Ma $\&$ Hovy \cite{ma2016end}                   &97.55
				\\ \hline
				Linear-chain NCRF       &97.52 \\
				RNN transducer   &97.50 \\
				NCRF transducer   &\bf 97.52 \\
				\bottomrule
			\end{tabular}
		\end{center}	
	\end{footnotesize}
\vspace{-0.1cm}
\end{table}

\begin{table}[t]
	\begin{footnotesize}
		\begin{center}
			\caption{Chunking results, trained only with  CoNLL-2000 data.}
			\label{table：chunking}
			\begin{tabular}{lll}
				\toprule
				\multicolumn{1}{l}{\bf Model}  &\multicolumn{1}{l}{$F_1\pm$ \bf std} &\multicolumn{1}{l}{$F_1$ \bf max} \\
				\midrule
				S$\phi$gaard $\&$ Goldberg \cite{sogaard2016deep}     &95.28 \\
				Hashimoto et al \cite{hashimoto2017joint}         &95.02 \\			
				Yang et al \cite{yang2017transfer}             &94.66 \\
				Peters et al \cite{peters2017semi}    &95.00 $\pm$ 0.08
				\\ \hline
				Linear-chain NCRF       &95.01 $\pm$ 0.12 &95.15 \\
				RNN transducer   &95.02 $\pm$ 0.11 &95.13 \\
				NCRF transducer   & \bf 95.14 $\pm$ 0.05 & \bf 95.23  \\
				\bottomrule
			\end{tabular}
		\end{center}
	\end{footnotesize}
\vspace{-0.4cm}
\end{table}

\begin{table}[t]
	\begin{footnotesize}
		\begin{center}
			\caption{English NER results, trained only with CoNLL-2003 data.}
			\label{table:English NER}
			\begin{tabular}{lll}
				\toprule
				\multicolumn{1}{l}{\bf Model}  &\multicolumn{1}{l}{ $F_1\pm$ \bf std} &\multicolumn{1}{l}{$F_1$ \bf max} \\
				\midrule
				Luo et al \cite{luo2015joint}            &89.90 \\
				Chiu $\&$ Nichols \cite{chiu2016named}         &90.91 $\pm$ 0.20 \\
				Lample et al \cite{lample2016neural}             &90.94 \\
				Ma $\&$ Hovy \cite{ma2016end}           &91.21 \\
				Yang et al \cite{yang2017transfer}          &91.20 \\
				Peters et al \cite{peters2017semi}         &90.87 $\pm$ 0.13 \\
				Liu et al \cite{liu2017empower}            &91.24 $\pm$ 0.12      &91.35
				\\ \hline
				Linear-chain NCRF       &91.11 $\pm$ 0.16   &91.30 \\
				RNN transducer   &91.02 $\pm$ 0.15 &91.23 \\
				NCRF transducer   &\bf91.40 $\pm$ 0.11 &\bf91.66 \\
				\bottomrule
			\end{tabular}
		\end{center}
	\end{footnotesize}
\vspace{-0.1cm}
\end{table}

\begin{table}[t]
	\begin{footnotesize}
		\begin{center}
			\caption{English NER results, trained only with CoNLL-2003 data but using other external resources.}
			\label{table:English NER with external resources}
			\vspace{-0.3cm}
			\begin{tabular}{llll}
				\toprule
				\multicolumn{1}{l}{\bf Model} &\multicolumn{1}{l}{\bf External resources} &\multicolumn{1}{l}{ $F_1\pm$ \bf std} 
				&\multicolumn{1}{l}{$F_1$ \bf max} \\
				\midrule
				Collobert et al \cite{collobert2011natural}         &gazetteers &89.59 \\
				Luo et al \cite{luo2015joint}             &entity linking &91.2\\
				Chiu $\&$ Nichols \cite{chiu2016named}             &gazetteers &91.62 \\
				Yang et al \cite{yang2017transfer}          &PTB-POS &91.26 \\
				Peters et al \cite{peters2018deep}     &ELMo &92.22 $\pm$ 0.10
				\\ \hline
				Linear-chain NCRF       &ELMo &92.23 $\pm$ 0.13 &92.51 \\
				RNN transducer   &ELMo &92.03 $\pm$ 0.21 &92.30\\
				NCRF transducer   &ELMo &\bf92.36 $\pm$ 0.20 &\bf92.61\\
				\bottomrule
			\end{tabular}
		\end{center}
	\end{footnotesize}
\vspace{-0.3cm}
\end{table}

\begin{table}[t]
	\begin{footnotesize}
		\begin{center}
			\caption{Dutch NER results, trained only with CoNLL-2002 data.}
			\label{table: Dutch NER}
			\begin{tabular}{lll}
				\toprule
				\multicolumn{1}{l}{\bf Model}  &\multicolumn{1}{l}{ $F_1\pm$ \bf std} &\multicolumn{1}{l}{$F_1$ \bf max} \\
				\midrule
				Nothman et al \cite{nothman2013learning}         &78.6 \\
				Gillick et al \cite{gillick2016multilingual}             &78.08 \\
				Lample et al \cite{lample2016neural}             &81.74
				\\ \hline
				Linear-chain CRF       &81.53 $\pm$ 0.31 &81.76\\
				RNN transducer   &81.59 $\pm$ 0.09 &81.70 \\
				NCRF transducer   &\bf81.84 $\pm$ 0.07 &\bf81.94\\
				\bottomrule
			\end{tabular}
		\end{center}
	\end{footnotesize}
\vspace{-0.4cm}
\end{table}

\section{Experimental results}
\label{sec:Results}

For each method, we perform five independent runs with  random initializations and report the mean and standard deviation. For comparison with other top-performance systems, we report the best results from the two potential designs - the design in Eq. (\ref{eq:design-a}) for POS tagging and chunking and the design in Eq. (\ref{eq:design-b}) for NER.

\subsection{Comparison of NCRF transducers with linear-chain NCRFs and RNN Transducers}
\label{ssec:Comparison with CRF and RNN Transducer}

For fair comparison of different methods, the same transcription network architecture is used in linear-chain NCRFs, RNN transducers and NCRF transducers in our experiments.

First, it can been seen from Table \ref{table:pos}, \ref{table：chunking}, \ref{table:English NER}, \ref{table:English NER with external resources} and \ref{table: Dutch NER} that 
NCRF transducers outperform linear-chain NCRFs for three out of the four sequence labeling tasks, which clearly shows the benefit of long-range dependency modeling.
Numerically, the improvements are: 0.13 mean $F_1$ in chunking, 0.29/0.13 mean $F_1$ in English NER (without/with ELMo), and 0.31 mean $F_1$ in Dutch NER.
Such equal performances in POS tagging suggests that long-range dependencies are not significant in POS tagging.

Second, NCRF transducers outperform RNN transducers across the four sequence labeling tasks. 
Numerically, the improvements are: 0.02 accuracy in POS tagging, 0.12 mean $F_1$ in chunking, 0.38/0.33 mean $F_1$ in English NER (without/with ELMo), 0.25 mean $F_1$ in Dutch NER.
This demonstrates the advantage of overcoming the label bias and exposure bias problems in NCRF transducers.
Note that to solve the exposure bias problem, scheduled sampling has been applied to training RNNs \cite{bengio2015scheduled}. We also train RNN transducers with scheduled sampling, but no better results are obtained.

\subsection{Comparison with prior results}
\label{Comparison with state-of-the-art results}

In the following, we compare the results obtained from NCRF transducers with prior results to show the significance of NCRF transducers, but noting that there are differences in network architectures and other experimental conditions.

\textbf{POS tagging.}
As shown in Table~\ref{table:pos}, NCRF transducer achieves highly competitive accuracy of 97.52, close to 97.55 obtained by Bi-LSTM-CNNs-CRF in \cite{ma2016end}.

\textbf{Chunking.}
As shown in Table~\ref{table：chunking}, NCRF transducer achieves 95.14 mean F1, representing an improvement of 0.14 mean $F_1$ over an advanced linear-chain NCRF \cite{peters2017semi}. 
The higher result of 95.28 \cite{sogaard2016deep} is obtained with deep bi-RNNs models.

\textbf{English NER.}
Table~\ref{table:English NER} and \ref{table:English NER with external resources} show the results without/with external resources respectively. 
In both cases, NCRF transducers produce new state-of-the-art results. When no external resources are used, NCRF transducer gives 91.40 mean $F_1$, outperforming the previous best result 91.24 (linear-chain NCRF) \cite{liu2017empower}. When augmented with ELMo \cite{peters2018deep}, NCRF transducer further achieves 92.36 mean $F_1$, better than the original ELMo result.

\textbf{Dutch NER.}
As shown in Table~\ref{table: Dutch NER}, NCRF transducer achieves 81.84 mean $F_1$, which is better than 81.74 from \cite{lample2016neural} (linear-chain NCRF), when using the same word embeddings in \cite{lample2016neural}. 
The results of 85.19 in \cite{yang2017transfer} and 82.84 in \cite{gillick2016multilingual} are obtained using different word embeddings with the help of transfer learning \cite{yang2017transfer} or multilingual data \cite{gillick2016multilingual}, so these two results cannot be directly compared to the results of NCRF transducers.

\section{Related Work}
\label{sec:related work}


Extending CRFs to model higher-order interactions than pairwise relationships between labels is an important issue for sequence labeling.
There are some prior studies, e.g. higher-order CRFs \cite{Chatzis2013TheIC}, semi-Markov CRFs \cite{Sarawagi2004SemiMarkovCR} and latent-dynamic CRFs \cite{morency2007latent}, but not using NNs.
Using NNs to enhance the modeling of long-range dependencies in CRFs is under-appreciated in the literature. 
A related work is structured prediction energy networks (SPENs) \cite{belanger2016structured}, which use
neural networks to define energy functions that potentially can capture long-range dependencies between structured outputs/labels.
SPENs depend on relaxing labels from discrete to continuous and use gradient descent for test-time inference, which is time-consuming. Training and inference with SPENs are still challenging, though with progress \cite{tu2018learning}.

Our work on developing (globally normalized) NCRF transducers is related to prior studies on developing globally normalized models \cite{andor2016globally,wiseman2016sequence}, which aim to overcome the label bias and exposure bias problems that locally normalized
models suffer from.
The beam search training scheme in our work is close to \cite{andor2016globally,wiseman2016sequence}, but the developed models are different.
The globally normalized model in \cite{andor2016globally} defines a CRF which uses feedforward neural networks to implement potentials but does not address long-range dependencies modeling in CRFs.
The work in \cite{wiseman2016sequence} extends the
seq2seq model by removing the final softmax in the RNN decoder to learn global sequence scores.

\section{Conclusion}
\label{sec:conclusion}

In this paper, we propose neural CRF transducers, which consists of two RNNs, one extracting features from observations and the other capturing (theoretically infinite) long-range dependencies between labels.
Experiment results show that NCRF transducers achieve consistent improvements over linear-chain NCRFs and RNN transducers across four sequence labeling tasks, and can improve state-of-the-art results.
An interesting future work is to apply NCRF transducers to those structured prediction tasks, e.g. speech recognition, where long-range dependencies between labels are more significant.


\bibliographystyle{IEEEbib}
\bibliography{strings,refs}

\end{document}